\documentclass[11pt]{article}
\usepackage{coling2020}
\usepackage{times}
\usepackage{url}
\usepackage{latexsym}
\usepackage{bold-extra}
\usepackage{hyperref}
\usepackage{caption}
\usepackage[ruled,vlined]{algorithm2e}
\usepackage[export]{adjustbox}
\usepackage{wrapfig}
\usepackage{mathtools}
\usepackage[table]{xcolor}
\SetKwComment{Comment}{$\triangleright$\ }{}
\usepackage[titletoc,title]{appendix}
\usepackage{etoolbox}
\apptocmd{\appendices}{\apptocmd{\thesection}{.}{}{}}{}{}

\def\@fnsymbol#1{\ensuremath{\ifcase#1\or \dagger\or \mathsection\or \ddagger\or
   *\or \mathparagraph\or \|\or **\or \dagger\dagger
   \or \ddagger\ddagger \else\@ctrerr\fi}}
\newcommand{\ssymbol}[1]{^{\@fnsymbol{#1}}}

\colingfinalcopy % Uncomment this line for the final submission

\title{Towards A Friendly Online Community:\\ An Unsupervised Style Transfer Framework for Profanity Redaction}

  \author{Minh Tran$\ssymbol{1}$, Yipeng Zhang$\ssymbol{2}$, Mohammad Soleymani$\ssymbol{1}$\\
  $\ssymbol{1}$Institute for Creative Technologies, University of Southern California, Los Angeles, CA, USA\\ 
  $\ssymbol{2}$University of Rochester, Rochester, NY, USA\\
  $\ssymbol{1}$\texttt{\{mtran,soleymani\}@ict.usc.edu}\\
  $\ssymbol{2}$\texttt{yzh232@u.rochester.edu}}

\date{}

\begin{document}
\maketitle
\begin{abstract}
%With the development of social media platforms, it is an urgent concern that offensive comments are rampant on the internet. In this work, we focus on the task of transforming offensive comments, which are comments that contains profanity or offensive languages, into non-offensive ones. We design a \textsc{Retrieve}, \textsc{Generate} and \textsc{Edit} unsupervised style transfer pipeline to regenerate the offensive comments in a word-restricted manner while maintaining a high level of fluency and preserving the content of the original text. We evaluate and extensively compare our method's performance with multiple prior style transfer models on both automatic metrics and human evaluations.
Offensive and abusive language is a pressing problem on social media platforms. In this work, we propose a method for transforming offensive comments, statements containing profanity or offensive language, into non-offensive ones. We design a \textsc{Retrieve}, \textsc{Generate} and \textsc{Edit} unsupervised style transfer pipeline to redact the offensive comments in a word-restricted manner while maintaining a high level of fluency and preserving the content of the original text. We extensively evaluate our method's performance and compare it to previous style transfer models using both automatic metrics and human evaluations. Experimental results show that our method outperforms other models on human evaluations and is the only approach that consistently performs well on all automatic evaluation metrics.
\end{abstract}

\section{Introduction}
\label{intro}

Despite the undeniably positive impact social media has on facilitating communication, it is also a medium that can be used for abusive behavior. Many social media platforms do not restrict the language users use, leading to an overflow of strong language that might not be appropriate for children \cite{duggan2014online,rieder2010no}. Verbal abuse and cyber-bullying is also a common problem on social media. Such phenomena are harmful to the victims, the online community, and in particular adolescents who are more susceptible and vulnerable in such situations \cite{patchin2010cyberbullying,pieschl2015beware}. To mitigate such problems, recent studies have focused on developing machine learning models for detecting hate speech \cite{davidson2017automated,xiang2012detecting,djuric2015hate,waseem2016hateful,chen2012detecting,xiang2012detecting,founta2019unified}. However, little progress has been made regarding the task of transforming hateful sentences into non-hateful ones, a potential next-step after detecting the hateful content. \newcite{dos2018fighting} propose an extension of a basic encoder-decoder architecture by including a collaborative classifier. To the best of our knowledge, this is the only approach dealing with abusive language redaction.

Unsupervised text style transfer is an important area in text generation that has recently received a lot of attention. Generally speaking, text style transfer is the task of rewriting sentences in a source style to a target style while preserving the original sentences as much as possible. In the context of the paper, we define a corpus to be stylistic if every sample in the corpus shares a common style. Most style transfer approaches are developed and validated on bi-stylistic datasets \cite{shen2017style,hu2017toward,li2018delete,prabhumoye2018style,tian2018structured,he2019probabilistic,wu2019mask}, which require stylistic features on both source and target samples. Some common bi-stylistic datasets for text style transfer are the (negative-positive) Yelp restaurant reviews \cite{shen2017style} \& Amazon product reviews \cite{he2016ups}, (democratic-republican) Political slant \cite{prabhumoye2018style}, (male-female) Gender \cite{reddy2016obfuscating} and (factual-romantic-humorous) Caption  \cite{gan2017stylenet}. Models training on these datasets are not normally suitable for being trained and validated on uni-stylistic datasets, where only the source or the target set is stylistic (\emph{e.g.}, offensive to normal text). Recently, \newcite{madaan-etal-2020-politeness} introduce a uni-stylistic Politeness dataset along with a tag-and-generate approach, in which a generator model learns style phrases from the target samples to fill in tagged positions (cannot be generalized to our case where the target sentences are not stylistic).

In this work, we propose a novel \textsc{Retrieve}, \textsc{Generate} and \textsc{Edit} framework to solve the task of transferring offensive sentences into non-offensive ones. For validation, we use three criteria for assessing the performance of our model, namely, content preservation, style transfer accuracy and fluency. We perform an extensive comparison with prior style transfer work on both objective and subjective ratings.

\section{Methodology}
\subsection{Problem Formulation}
Given a vocabulary of restricted words $V_r$ and a corpus of labeled sentences $\mathcal{D}=\{(x_1, l_1),\dots, (x_n, l_n)\}$ where $x_i$ is a sentence and $l_i=$ ``offensive" if there exists an offensive word $v_i$ ($v_i \in V_r$) in  $x_i$, otherwise $l_i=$ ``non-offensive". For $(x_i, l_i)$ where $l_i=$ ``offensive", we re-generate $x_i^*$ such that it does not contain any words from $V_r$, preserves as much content from $x_i$ as possible, and is grammatical and fluent. Unlike \newcite{dos2018fighting}, who handle general hateful and offensive content detected by \newcite{davidson2017automated}'s offensive language and hate speech classifier, we focus our work on profanity removal.

\begin{figure}[ht]
    \centering
    \includegraphics[width=\textwidth]{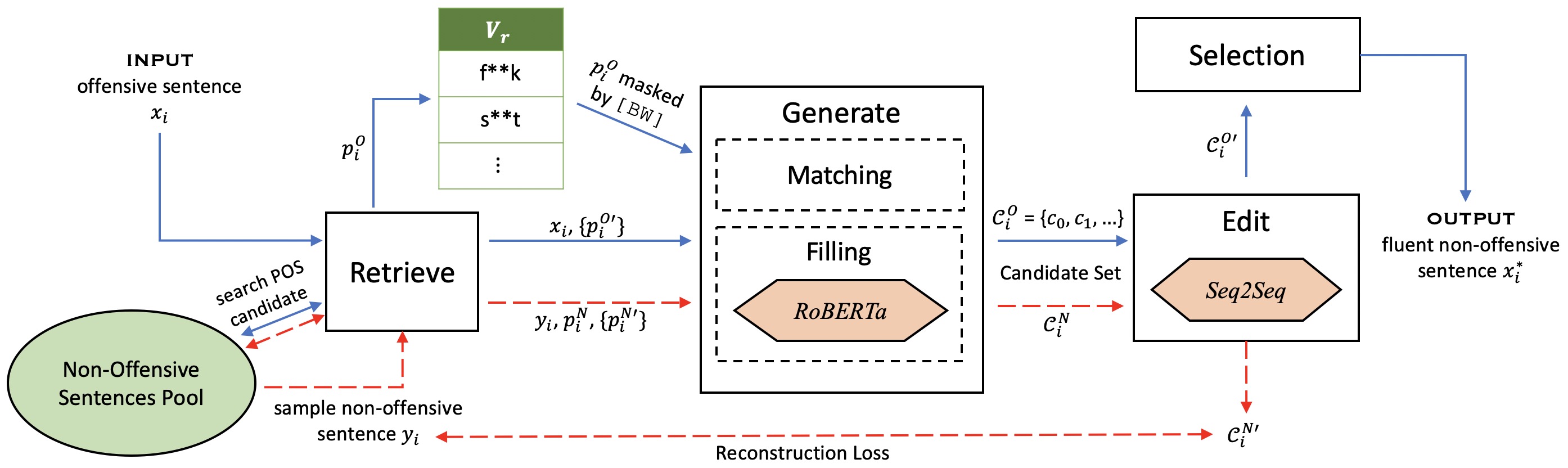}
    \caption{Overview of our \textsc{Retrieve-Generate-Edit} framework. The \textcolor{red}{dotted red} arrows denote the steps for training the sequence-to-sequence model while the \textcolor{blue}{solid blue} ones denote the steps taken during inference. We use superscripts $O$ (offensive) and $N$ (non-offensive) to differentiate the variables.}
    \label{fig:model}
\end{figure}

\subsection{Data Collection}

\iffalse
\begin{wrapfigure}{R}{0.5\textwidth}
\begin{minipage}{\linewidth}
\begin{algorithm}[H]
\SetKwInOut{Input}{Input}
\SetKwInOut{Output}{Output}
\SetAlgoLined
\DontPrintSemicolon
\Input{$x_i$, $p_i'$, $V_r$, $\mathcal{T}_{i}$ - set of unique POS tokens in $p_i$, $\mathcal{T}_{i}'$ - set of unique POS tokens in $p_i'$, $\mathcal{F}$ - pretrained mask-filling model.}  
\Output{Set of candidate sentences $\mathcal{C}_i$.}
\textbf{Definition:} $P \coloneqq$ Number of permutations. \\
$\mathcal{T}_{shared}$ $\leftarrow$ $\mathcal{T}_{i}$ $\cap$ $\mathcal{T}_{i}'$ \\
$c_0 \leftarrow [\texttt{MASK}]_1 [\texttt{MASK}]_2 \dots [\texttt{MASK}]_{|p_i'|}$\\
$\mathcal{C}_i \leftarrow \{c_0\}$\\

\ForEach{token $t_k$ in $\mathcal{T}_{shared}$}{
    $\mathcal{W}_k \leftarrow$ set of words in $x_i$ tagged with $t_k$\\
    $\mathcal{S}_k \leftarrow$ list of $t_k$'s positions in $p_i'$\\
    $\mathcal{A}_k \leftarrow$ list of possible assignments of words in $\mathcal{W}_k$ to positions $\mathcal{S}_k$ \Comment*[r]{$\mathcal{O}(max(P^{|\mathcal{W}_k|}_{|\mathcal{S}_k|}, P^{|\mathcal{S}_k|}_{|\mathcal{W}_k|}))$}
    \ForEach{candidate $c_j$ in $\mathcal{C}_i$}{
        $\mathcal{C}_i$.remove($c_{j}$)\\
        \ForEach{assignment $a_l$ in $\mathcal{A}_k$}{
        $c_{jl} \leftarrow $\texttt{MATCH}($a_l$, $c_j$)\\
        $\mathcal{C}_i$.add($c_{jl}$)
        }
    }
}
\Return $\{\mathcal{F}(c_j)\}_{j=0, 1, \dots, |\mathcal{C}_i|}$
 \caption{Candidate set generation.}
 \label{alg:gen}
\end{algorithm}
\end{minipage}
\end{wrapfigure}

\fi

We construct the list of 1,580 restricted words $V_r$ from various sources \footnote{\url{https://www.noswearing.com/dictionary}}\footnote{\url{https://www.cs.cmu.edu/~biglou/resources/bad-words.txt}}. For Corpus $\mathcal{D}$, we extract a total of 12M comments from 2 highly controversial subreddits (6M from each): \texttt{r/The\_Donald} and \texttt{r/politics} from January 2019 to December 2019 using \textit{BigQuery}\footnote{\url{https://cloud.google.com/bigquery}}. We extract sentences that have between 5 and 20 words from the comments. We further remove sentences containing URL, number, email, emoticon, date and time using the \textit{Ekphrasis} text normalization tool \cite{baziotis2017datastories}. The remaining sentences are then labeled as either ``offensive" or ``non-offensive", as defined, resulting in 350K ``offensive" sentences and 7M ``non-offensive" sentences. 

\subsection{Framework}
As shown in Figure~\ref{fig:model}, our \textsc{Retrieve}, \textsc{Generate} and \textsc{Edit} framework first retrieves possible Part-of-Speech (POS) tagging sequences, which are then used as the templates for generating candidates in the \textsc{Generate} module and corrected by the \textsc{Edit} module.

\paragraph{Retrieve} We first perform POS tagging on both the labelled 350K offensive and 7M non-offensive comments using the \textit{Stanza} POS tagger \cite{qi2020stanza}. We replace the POS tags of the offensive terms in $V_r$ with a \texttt{[BW]} token. Then, given an offensive sentence $x_i$ and its POS sequence $p_i$, we use the \textit{Lucene} search engine\footnote{\url{https://lucene.apache.org/core/}}(TF-IDF based) to find the set of 10 most similar POS sequences $\{p_i'\}$ that belong to sentences in the non-offensive set.

\paragraph{Generate} After getting $x_i$, $p_i$ and $\{p_i'\}$, the \textsc{Generate} module creates a set of sentences $\mathcal{C}_i$ containing no offensive words. The module achieves this by ``matching" words in $x_i$ into possible positions in each $p_i'$ to generate new sentences. The positions that are unable to be matched are ``filled" by a pretrained language model. The pseudocode for the algorithm can be found in Algorithm \ref{alg:gen}.

{\centering
\begin{minipage}{.8\linewidth}
\begin{algorithm}[H]
\SetKwInOut{Input}{Input}
\SetKwInOut{Output}{Output}
\SetAlgoLined
\DontPrintSemicolon
\Input{$x_i$, $p_i$, $p_i'$, $V_r$ \\$\mathcal{T}_{i}$ - set of unique POS tokens in $p_i$ \\$\mathcal{T}_{i}'$ - set of unique POS tokens in $p_i'$\\ $\mathcal{F}$ - pretrained mask-filling model}  
\Output{Set of candidate sentences $\mathcal{C}_i$.}
\textbf{Definition:} $P^n_k \coloneqq$ Value of the k-permutations of n. \\
$\mathcal{T}_{shared}$ $\leftarrow$ $\mathcal{T}_{i}$ $\cap$ $\mathcal{T}_{i}'$ \\
$c_0 \leftarrow [\texttt{MASK}]_1 [\texttt{MASK}]_2 \dots [\texttt{MASK}]_{|p_i'|}$\\
$\mathcal{C}_i \leftarrow \{c_0\}$\\

\ForEach{token $t_k$ in $\mathcal{T}_{shared}$}{
    $\mathcal{W}_k \leftarrow$ set of words in $x_i$ tagged with $t_k$\\
    $\mathcal{S}_k \leftarrow$ list of $t_k$'s positions in $p_i'$\\
    $\mathcal{A}_k \leftarrow$ list of possible assignments of words in $\mathcal{W}_k$ to positions $\mathcal{S}_k$ \Comment*[r]{$\mathcal{O}(max(P^{|\mathcal{W}_k|}_{|\mathcal{S}_k|}, P^{|\mathcal{S}_k|}_{|\mathcal{W}_k|}))$}
    \ForEach{candidate $c_j$ in $\mathcal{C}_i$}{
        $\mathcal{C}_i$.remove($c_{j}$)\\
        \ForEach{assignment $a$ in $\mathcal{A}_k$}{
        $c_{j}' \leftarrow $\texttt{ASSIGN}($a$, $c_j$)\\
        $\mathcal{C}_i$.add($c_{j}'$)
        }
    }
}
\Return $\{\mathcal{F}(c_j)\}_{j=0, 1, \dots, |\mathcal{C}_i|}$
 \caption{Candidate set generation.}
 \label{alg:gen}
\end{algorithm}
\end{minipage}
\par
}

\begin{itemize}
    \item \textbf{Matching} 
For each $p_i'$, we first create a set $\mathcal{T}_{shared}$ of unique shared tokens in $p_i$ and $p_i'$. We initialize sentence $c_0$ of length $|p_i'|$ filled with \texttt{[MASK]} tokens to store the sentence generated according to $p_i'$. For a token $t_k$ in $\mathcal{T}_{shared}$, we try to fill all its corresponding positions in $c_0$ using words in $x_i$ that are tagged with $t_k$. Suppose there are $N$ words and $M$ positions, then there are at most $max(\frac{N!}{(N-M)!}, \frac{M!}{(M-N)!})$ possible permutations. We find the number to be 9.42 on average for 5K randomly sampled offensive sentences. We add each newly generated sentence $c_{j}'$ into $\mathcal{C}_i$ and repeat for each $t_k$ on all sentences in $\mathcal{C}_i$ until all their masked positions correspond to tokens not in $\mathcal{T}_{shared}$.
    \item \textbf{Filling}
For each resulting candidate sentence in $\mathcal{C}_i$, we use the pretrained RoBERTa-base model \cite{liu2019roberta} to fill in remaining \texttt{[MASK]} tokens. To enhance content preservation, we insert the original sentence $x_i$ before each of the generated sentences with a \texttt{[SEP]} token in between. We replace each \texttt{[SEP]} token with the most probable word predicted by RoBERTa that is not in \texttt{$V_r$}. The unmasked sentences after \texttt{[SEP]} are the outputs of the \textsc{Generate} module. 
\end{itemize}

\paragraph{Edit} We use an \textsc{Edit} module to correct the problems of the output sentences from the \textsc{Generate} module, mostly related to wrong word orderings due to the permutation generation in the \textsc{Matching} step or low fluency due to a bad retrieved POS sequence from the \textsc{Retrieve} module. We first randomly sample 60K English-only non-offensive sentences and apply the \textsc{Retrieve} and \textsc{Generate} modules on the chosen sentences (dotted red arrows in Figure~\ref{fig:model}). In the \textsc{Retrieve} module, we retrieve POS sequences $\{{p^N_i}'\}$ from the non-offensive set and drop the first retrieved sequence, which is the original query sequence $y_i$ itself. We then form a parallel corpus using the generated candidates $\mathcal{C}_i^N$ as the source dataset while having the original non-offensive sentences as the target dataset, resulting in 780K source-target pairs. In this study, we finetune the pretrained T5-small model \cite{raffel2019exploring} as our editing sequence-to-sequence model using the generated parallel corpus. We call the edited candidate set $\mathcal{C}_i'$.

\paragraph{Selection} We add a \textsc{Selection} module to select the candidate of highest quality $x_i^*$ from $\mathcal{C}_i'$. We first remove any candidate with words in $V_r$. Then, each generated candidate is assigned a content preservation score (BLEU score \cite{papineni2002bleu} between the source and the candidate sentences) and a fluency score (perplexity estimated by the pretrained GPT-2 model with 117M parameters\footnote{\url{https://huggingface.co/gpt2}} \cite{radford2019language}). The content preservation and fluency scores are then normalized to $[0,1]$ by \textit{MinMaxScaler}. The candidate with the highest sum of content preservation and fluency scores is chosen.

\section{Experimental Results}
\subsection{Baselines}
We compare our framework (\texttt{R+G+S} and \texttt{R+G+E+S})\footnote{\textsc{Retrieve}, \textsc{Generate}, [\textsc{Edit}] and \textsc{Selection}. The \textsc{Edit} module can be skipped.} against 8 existing style transfer methods. These methods are: cross-alignment \texttt{CA} \cite{shen2017style}, back-translation \texttt{BT} \cite{prabhumoye2018style}, delete-only \texttt{DL} and delete-retrieve-generate \texttt{DRG} \cite{li2018delete}, mask-and-infill \texttt{MLM} \cite{wu2019mask}, auto-encoder with POS information preservation constraint \texttt{AEC} \cite{tian2018structured}, deep latent sequence model \texttt{DLS} \cite{he2019probabilistic} and the tag-and-generate model \texttt{TG} \cite{madaan-etal-2020-politeness}. We also compare our method with the removal approach \texttt{REM}, which simply removes offensive terms from sentences.

For all baselines methods, we replicate the experimental setups described in their papers. Since some of the baseline models' performance are susceptible to unbalanced classes during training  \cite{li2018delete,wu2019mask,tian2018structured}, we subsample the non-offensive sentences from 7M to 350K sentences, resulting in a balanced dataset. We then split the offensive and non-offensive datasets into train (320K), validation (25K)  and test (5K) sets. Implementation details can be found in Appendix~\ref{app:implementation}

% \iffalse
% \subsection{Implementation Details}
% For the \textsc{Retrieve} module, we collect the 10 best retrieved outputs if \texttt{$x_i$} is labelled as offensive and 5 best retrieved outputs if \texttt{$x_i$} is labelled as non-offensive (creating data for the \textsc{Edit} module). For generating data in the \textsc{Edit} module, we random sample a dataset of around 60K English-only sentences from the PST dataset and use these sentences as inputs for the \textsc{Retrieve} and \textsc{Generate} modules, resulting in a dataset of 780K source-target pairs of sentences. 
% \fi

% We use a 6-layered Transformer, each layer having $8$ attention heads with $512$ dimensional embedding layer. We train the model with a drop-out rate of $0.1$, the Adam optimizer [cite], a training batch size of $2048$ and a learning rate of $0.001$ for $5$ epochs. We finetune the T5-small model with a learning rate of $1e^{-4}$, the Adam optimizer and a training batch size of $256$ for $3$ epochs. For both the Transformer and T5 models, we use a beam size of 5 and set the max length of input and output sequences to be 30.

\subsection{Evaluations}
\begin{wrapfigure}{R}{0.56\textwidth}
\begin{minipage}{\linewidth}
% \begin{table}[h]
% \begin{center}
% \begin{tabular}{|l|c|c|c|c|c|c|}
\begin{tabular}{|l|lllll|}
\hline \bf Model & \bf BL $\uparrow$ & \bf RG $\uparrow$ & \bf MT $\uparrow$ & \bf Acc. $\uparrow$ & \bf PPL $\downarrow$ \\
\hline
\iffalse
POL & 66.1 & 76.3 & 45.4 & 23.4 & 3378.5  \\
DL & 51.8 & 63.4 & 30.1 & 56.8 & 811.0  \\
DRG & 47.9 & 59.6 & 28.3 & 57.2 & 1113.4  \\
\hline
\fi
CA & \cellcolor{red!20}18.3 & \cellcolor{red!20}36.2 & \cellcolor{red!20}11.9  & \cellcolor{red!40}65.0 & \cellcolor{red!10}747.7 \\
MLM & \cellcolor{green!20}49.7 & \cellcolor{green!20}63.3 & \cellcolor{green!20}40.8 & \cellcolor{red!20}65.5 & \cellcolor{red!20}798.6 \\
AEC & 46.7 & 56.3 & 25.9 & \cellcolor{red!10}90.2 & \cellcolor{red!40}3470.6  \\
BT & \cellcolor{red!40}8.5 & \cellcolor{red!40}21.3 & \cellcolor{red!40}9.3 & 95.2 & \cellcolor{green!10}488.5  \\
DLS & \cellcolor{red!10} 30.9 & \cellcolor{red!10}48.8 & \cellcolor{red!10}17.9  & \cellcolor{green!10}99.1 & \cellcolor{green!40}445.9  \\
\hline
R+G+S & \cellcolor{green!40}51.8 & \cellcolor{green!40}67.7 & \cellcolor{green!40}41.5 & \cellcolor{green!40}100.0 & 674.9  \\
%R+G+Tr+S & 25.1 & 36.5 & 21.3 & 100.0 & 451.1  \\
R+G+E+S & \cellcolor{green!10}47.4 & \cellcolor{green!10}57.7 & \cellcolor{green!10}33.9 & \cellcolor{green!20}99.6 & \cellcolor{green!20}448.7  \\
\hline
REM & 81.3 & 87.9 & 49.0 & 100.0 & 1259.8  \\
\hline
\end{tabular}
% \end{center}
\captionof{table}{\label{font-table} Automatic evaluation results. For each metric, we mark the 3 best/worst-performing models in \textcolor{green}{green}/\textcolor{red}{red}. The average perplexity of the original sentences is $458.1$.}
\label{automatic_evaluation}
% \end{table} 
\end{minipage}
\end{wrapfigure}

\paragraph{Automatic Evaluations} Following most prior studies on text style transfer, we use 3 criteria to evaluate the generated outputs of the models: content preservation, style transfer accuracy and fluency. For content preservation, we report the BLEU-self~(BL) \cite{papineni2002bleu}, ROUGE~(RG) \cite{lin2004rouge} and METEOR~(MT) \cite{denkowski2011meteor}. We calculate the style transfer accuracy~(Acc.) as the percentage of generated sentences not containing any words in $V_r$. For fluency, we use the average perplexity~(PPL) of generated sentences calculated by the pretrained GPT-2 model \cite{radford2019language}.

We show the performances of methods that have at least 60\% accuracy in Table~\ref{automatic_evaluation}, while reporting the remaining ones in Appendix~\ref{app:results} Our models are the only ones that consistently perform among the top in all 3 criteria. %On the other hand, \texttt{BT} has both a high accuracy and a good fluency rating but fails to maintain the original content. 
The perplexity of \texttt{R+G+E+S} is lower than the perplexity of \texttt{R+G+S} by $226$ points, suggesting the effectiveness of the trained sequence-to-sequence model to edit the output candidates from the \textsc{Generate} module.

 Although we do not compare the performance of our framework with \cite{dos2018fighting}, we use the same set of evaluation metrics reported in their work. On a training dataset of size 224K offensive sentences and 7M non-offensive Reddit sentences, \newcite{dos2018fighting} report a content preservation score, proposed by \newcite{fu2018style}, of 0.933, a style transfer accuracy of 99.54\% and a worse perplexity than \texttt{CA}'s outputs. For reference, our best performing model, \texttt{R+G+E+S}, achieves a \newcite{fu2018style}'s content preservation score of $0.965$, a style transfer accuracy of 99.6\% and a better perplexity than \texttt{CA}.

\begin{wrapfigure}{R}{0.51\textwidth}
\begin{minipage}{\linewidth}
\begin{tabular}{|l|c|c|c|c|}
\hline \bf Model & \bf CP $\uparrow$ & \bf Gra. $\uparrow$ & \bf Acc. $\uparrow$ &\bf Succ. $\uparrow$\\ \hline
DLS & 1.947 & 4.037 & 99\% & 7\% \\
MLM & 3.157 & \textbf{4.383} & 73\% & 18\% \\
R+G+S & \textbf{3.650} & 3.840 & \textbf{100}\% & 40\% \\
R+G+E+S & 3.567 & 4.077 & \textbf{100}\% & \textbf{46}\% \\
\hline
\end{tabular}
\captionof{table}{\label{font-table} Human evaluation results. }
\label{human_eval}
\end{minipage}
\end{wrapfigure}

\paragraph{Human Evaluations} We ask 3 unbiased human judges to rate the outputs of our models, as well as \texttt{MLM} and \texttt{DLS}, which are the 4 best performing models according to the automatic evaluation metrics. Following \newcite{li2018delete}, the annotators judge the generated sentences on content preservation (CP) and grammaticality (Gra.) on a scale from 1 to 5. From 5K offensive sentences in the test set, we randomly sample 100 offensive sentences and ask the annotators to rate the generated outputs of the 4 models on these chosen sentences. We report the style transfer success rate (Succ.) for each method, which is calculated as the number of sentences that do not contain any words from $V_r$ and receive an average CP and Gra. scores of at least 4. Table~\ref{human_eval} shows the results of the manual evaluations, which demonstrates a significantly higher Succ. score of \texttt{R+G+S} and \texttt{R+G+E+S} in comparison with previously published models. 
%We provide the mean of the ratings on human annotated criteria (i.e. CP and Gra.). 
Some generated samples of the 4 methods are available in Appendix~\ref{app:samples}

\section{Conclusion}
In this paper, we propose a novel \textsc{Retrieve}, \textsc{Generate} and \textsc{Edit} text style transfer framework that redacts offensive comments on social media in a word-restricted manner. The experimental results on both automatic metrics and manual evaluations demonstrate the strong performance of our method over prior models for the given task. For future work, we envision the possibility of extending the framework by automatically detecting the restricted vocabulary set \texttt{$V_r$}. Such ability would enable the framework to be a robust style transfer method that is applicable to both uni-stylistic and bi-stylistic datasets.

\section*{Acknowledgements}
Research was in-part sponsored by the Army Research Office and was accomplished under Cooperative Agreement Number W911NF-20-2-0053. The views and conclusions contained in this document are those of the authors and should not be interpreted as representing the official policies, either expressed or implied, of the Army Research Office or the U.S. Government. The U.S. Government is authorized to reproduce and distribute reprints for Government purposes notwithstanding any copyright notation herein.

% \section{Translation of non-English Terms}

% It is also advised to supplement non-English characters and terms
% with appropriate transliterations and/or translations
% since not all readers understand all such characters and terms.
% Inline transliteration or translation can be represented in
% the order of: original-form transliteration ``translation''.

% \section{Length of Submission}
% \label{sec:length}

% The maximum submission length is 9 pages (A4) of content for long papers and 4 pages (A4) of content for short papers, 
% plus an unlimited number of pages for
% references (for both long and short papers). 
% Authors of accepted papers will be given additional space in
% the camera-ready version to reflect space needed for changes stemming
% from reviewers comments.

% Papers that do not
% conform to the specified length and formatting requirements may be
% rejected without review.

% \iffalse
% For papers accepted to the main conference, we will invite authors to provide a translation 
% of the title and abstract and a 1-2 page synopsis of the paper in a second 
% language of the authors' choice. Appropriate languages include but are not 
% limited to authors' native languages, languages spoken in the authors' place 
% of affiliation, and languages that are the focus of the research presented.
% \fi

% \section*{Acknowledgements}

% The acknowledgements should go immediately before the references.  Do
% not number the acknowledgements section. Do not include this section
% when submitting your paper for review.

% include your own bib file like this:
\bibliographystyle{coling}
\bibliography{coling2020}

\begin{appendices}

\section{Implementation Details}

\label{app:implementation}
We finetune the T5-small model in the \textsc{Edit} module with a learning rate of $1e^{-4}$, the Adam optimizer, cross-entropy loss function and a training batch size of $256$ for $3$ epochs. We set the max length of input/output sequences to be 30 and the beam size to be 5.

\setcounter{table}{0}

\section{Full Automatic Evaluation Results}
\label{app:results}
\begin{table}[h]
\begin{center}
% \begin{tabular}{|l|c|c|c|c|c|c|}
\begin{tabular}{|l|l|l|l|l|l|l|}
\hline \bf Model & \bf BL $\uparrow$ & \bf RG $\uparrow$ & \bf MT $\uparrow$ & \bf FuCP $\uparrow$ & \bf Acc. $\uparrow$ & \bf PPL $\downarrow$ \\
\hline
TG \cite{madaan-etal-2020-politeness} & 66.1 & 76.3 & 45.4 & 0.960 & 23.4 & 3378.5  \\
DL \cite{li2018delete} & 51.8 & 63.4 & 30.1 & 0.931 & 56.8 & 811.0  \\
DRG \cite{li2018delete} & 47.9 & 59.6 & 28.3 & 0.927 & 57.2 & 1113.4  \\
CA \cite{shen2017style} & 18.3 & 36.2 & 11.9  & 0.907 & 65.0 & 747.7 \\
MLM \cite{wu2019mask} & 49.7 & 63.3 & 40.8 & 0.983 & 65.5 & 798.6 \\
AEC \cite{tian2018structured} & 46.7 & 56.3 & 25.9 & 0.912 & 90.2 & 3470.6  \\
BT \cite{prabhumoye2018style} & 8.5 & 21.3 & 9.3 & 0.900 & 95.2 & 488.5  \\
DLS \cite{he2019probabilistic} & 30.9 & 48.8 & 17.9 & 0.915 & 99.1 & 445.9  \\
\hline
R+G+S (Ours) & 51.8 & 67.7 & 41.5 & 0.977 & 100.0 & 674.9  \\
%R+G+Tr+S & 25.1 & 36.5 & 21.3 & 100.0 & 451.1  \\
R+G+E+S (Ours) & 47.4 & 57.7 & 33.9 & 0.965 & 99.6 & 448.7  \\
\hline
REM (remove only) & 81.3 & 87.9 & 49.0 & 0.986 & 100.0 & 1259.8  \\
\hline
\end{tabular}
% \end{center}
\captionof{table}{\label{font-table} Complete Automatic evaluation results. \textbf{FuCP} refers to the content preservation metric proposed by \protect\newcite{fu2018style}.}
\label{automatic_evaluation}
\end{center}
\end{table} 

\section{Example Outputs}
\label{app:samples}
\begin{table}[h]
\begin{center}
\begin{tabular}{|l|rl|}
\hline \bf Model & \bf Generated Output  &\\ \hline
\textbf{Original} & reap what you sow \textcolor{red}{a*s} clowns. &\\
DLS &  except what you believe us. &\\
MLM & this is what you sow and amazing amazing. & \\
R+G+S & reap what you sow clowns. &\\
R+G+E+S & reap what you sow. &\\
\hline
\textbf{Original} & the benghazi \textcolor{red}{b*tch} is going down. &\\
DLS &  the president cruz is going down. &\\
MLM & the benghazi wall is going down. & \\
R+G+S & the benghazi is down going down. &\\
R+G+E+S & the benghazi girl is going down. &\\
\hline
\textbf{Original} & put your head back up your \textcolor{red}{a*s} and stfu. &\\
DLS &  	put your head back up your own and then. &\\
MLM & stuff your smart back up your \textcolor{red}{a*s} and amazing great. & \\
R+G+S & put your head back up your back, and stfu. &\\
R+G+E+S & stfu and put your head back up. &\\
\hline
\textbf{Original} & the \textcolor{red}{w*ore} of babylon speaks. &\\
DLS &  the state of least run. &\\
MLM & the house of babylon speaks. & \\
R+G+S & babylon speaks the woman of babylon speaks. &\\
R+G+E+S & the babylon speaks of it. &\\
\hline
\textbf{Original} & you obviously talking through your \textcolor{red}{b*tt} because you lack all sense of having a brain. &\\
DLS &  you obviously talking on your own words because you want a lot of free. &\\
MLM & you keep talking through your teeth because you lost all sense of having a brain. & \\
R+G+S & you're obviously talking all sense having through you lack all sense of a brain. &\\
R+G+E+S & you obviously lack all sense of having a brain through your mouth when talking. &\\
% \hline
%  \textbf{Original} & fox is gonna mess around and let that alligator mouth overload that tweety bird \textcolor{red}{a*s}. &\\
%  DLS &  trump is trying to show and let that failed that police. &\\
%  MLM & fox is gonna love around and let that alligator mouth overload that tweety bird \textcolor{red}{a*s}. & \\
%  R+G+S & nobody is gonna mess around and let that alligator mouth overload. &\\
%  R+G+E+S & fox is gonna mess around and let that alligator overload him with the. &\\
%  \hline
%  \textbf{Original} & \textcolor{red}{h*ll} of a ride the past year. &\\
%  DLS &  obama of a bit of the year. &\\
%  MLM & amazing of a ride the past year. & \\
%  R+G+S & a ride of the past year. &\\
%  R+G+E+S & a ride of the past year is. &\\
\hline
\textbf{Original} & no one gives a \textcolor{red}{d*mn} about what your platform is because it has no merit. &\\
DLS &  no one gives a about what your country is because it is no longer. &\\
MLM & no one gives a flip about what your platform is because it has no merit. & \\
R+G+S & one cares about what one is doing because it has no merit. &\\
R+G+E+S & no one cares about what he is doing because it has no merit for. &\\
\hline
\textbf{Original} & i have no sympathy for that \textcolor{red}{b*tch} and i never will. &\\
DLS &  i have no idea for that it would never will be. &\\
MLM & i have tremendous sympathy for that \textcolor{red}{b*tch} and i always will. & \\
R+G+S & i have no sympathy for i will and never. &\\
R+G+E+S & i have no sympathy for you and will never. &\\
\hline
\textbf{Original} & war is \textcolor{red}{h*ll} and he deserves it. &\\
DLS &  war is cruz and he did it. &\\
MLM & war is real and he knows it. & \\
R+G+S & It is and he deserves it. &\\
R+G+E+S & it is war and he deserves it. &\\
\hline
\end{tabular}
\end{center}
\caption{\label{font-table} Example outputs from our framework, \texttt{DLS} and \texttt{MLM} }
\end{table}
\end{appendices}

\end{document}